\documentclass[10pt,twocolumn,letterpaper]{article}

\usepackage[pagenumbers]{cvpr} 

\usepackage{graphicx}
\usepackage{amsmath}
\usepackage{amssymb}
\usepackage{booktabs}
\usepackage{float}
\usepackage{multirow}
\usepackage{enumitem}
\usepackage{xcolor}
\usepackage{mathtools}
\usepackage{soul}

\DeclareMathOperator*{\argmin}{arg\,min}
\newif\ifcomments
\newif\ifedits

\commentsfalse
 \editsfalse

\ifcomments
\newcommand{\comments}[1]{#1}
\else
\newcommand{\comments}[1]{}
\fi

\ifedits
\newcommand{\edits}[1]{\textcolor{red}{\bf [edit: #1]}}
\else
\newcommand{\edits}[1]{#1}
\fi

\newcommand{\vivek}[1]{\comments{\textcolor{purple}{\bf [vivek: #1]}}}
\definecolor{muchlater}{rgb}{0.7,1,.7}

\newcommand{\removed}[1]{\comments{\textcolor{lightgray}{\st{#1}}}}

\newcommand{\fct}{\textsc{Fct}}
\usepackage[pagebackref,breaklinks,colorlinks]{hyperref}

\usepackage[capitalize]{cleveref}
\crefname{section}{Sec.}{Secs.}
\Crefname{section}{Section}{Sections}
\Crefname{table}{Table}{Tables}
\crefname{table}{Tab.}{Tabs.}

\def\cvprPaperID{9822} 
\def\confName{CVPR}
\def\confYear{2022}
\newcommand{\info}{\emph{side-information}}

\makeatletter
\renewcommand*{\@fnsymbol}[1]{\ifcase#1\or *\else$\dagger$\fi}
\makeatother

\begin{document}

\title{Forward Compatible Training for Large-Scale Embedding Retrieval Systems}

\author{Vivek Ramanujan\thanks{Corr. \href{mailto:ramanv@cs.washington.edu}{ramanv@cs.washington.edu} \& \href{mailto:mpouransari@apple.com}{mpouransari@apple.com}.}\\
University of Washington\thanks{Work completed during internship at Apple.}\\
\and
Pavan Kumar Anasosalu Vasu\\
Apple\\
\and
Ali Farhadi\\
Apple\\
\and
Oncel Tuzel\\
Apple\\
\and
Hadi Pouransari\textsuperscript{*}\\
Apple\\
}

\maketitle

\begin{abstract}
In visual retrieval systems, updating the embedding model requires recomputing features for every piece of data. This expensive process is referred to as backfilling. Recently, the idea of backward compatible training (BCT) was proposed. To avoid the cost of backfilling, BCT modifies training of the new model to make its representations compatible with those of the old model. However, BCT can significantly hinder the performance of the new model. In this work, we propose a new learning paradigm for representation learning: forward compatible training (\fct{}). In \fct{}, when the old model is trained, we also prepare for a future unknown version of the model. We propose learning side-information, an auxiliary feature for each sample which facilitates future updates of the model. To develop a powerful and flexible framework for model compatibility, we combine side-information with a forward transformation from old to new embeddings. Training of the new model is not modified, hence, its accuracy is not degraded. We demonstrate significant retrieval accuracy improvement compared to BCT for various datasets: ImageNet-1k (+$18.1\%$), Places-365 (+$5.4\%$), and VGG-Face2 ($+8.3\%$). \fct{} obtains model compatibility when the new and old models are trained across different datasets, losses, and architectures.\footnote{Code available at \url{https://github.com/apple/ml-fct}.}
\end{abstract}


\section{Introduction}\label{sec:intro}

\begin{figure}[t]
    \centering
    \includegraphics[width=1.0\columnwidth]{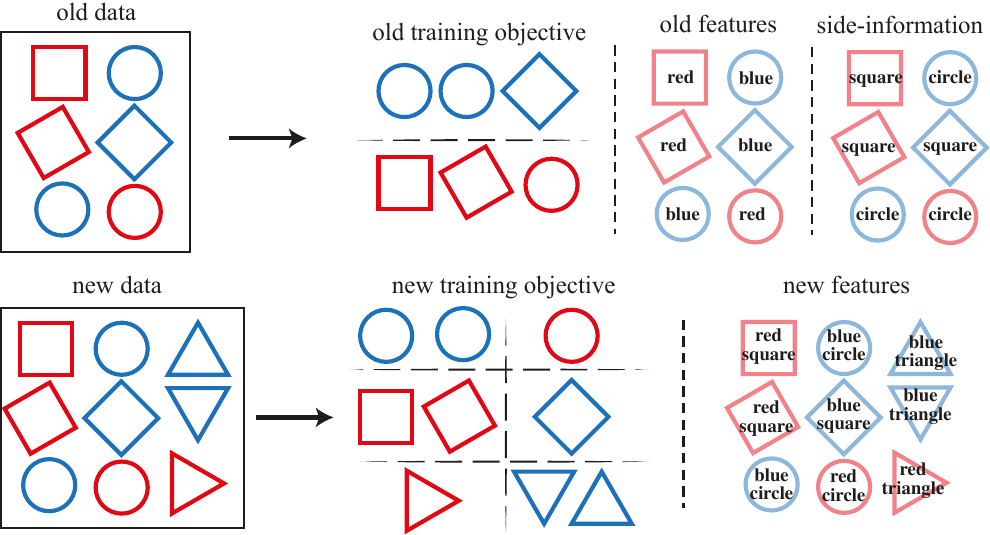}
    \caption{The old \edits{training objective} \textbf{(top)} requires discrimination between \emph{colors} of objects. The new \edits{training objective} \textbf{(bottom)} further requires discrimination between \emph{shapes} of objects. A model which learns the minimum description \edits{length} feature (\emph{color} \textbf{top middle}) of the old \edits{training objective} will be unable to complete the new \edits{training objective}. This demonstrates how side-information \textbf{(top right)} stores information which can be useful for future \edits{training objectives} but not necessarily the present \edits{objective}.}\vspace{-15pt}
    \label{fig:toy}
\end{figure}

Modern representation learning systems for vision use deep neural networks to embed very high-dimensional images into low-dimensional subspaces. These embeddings are multi-purpose representations that can be used for downstream tasks such as recognition, retrieval, and detection. In large systems in the wild, images are constantly being added in a highly distributed manner. Updating the models in these kinds of systems is challenging. The new model architecture and \edits{training objective} may be completely disjoint from that of the old model, and therefore its embeddings are \emph{incompatible} with downstream tasks reliant on the old embeddings. 

The naive solution to feature incompatibility, recomputing embeddings (\emph{backfilling}) for every image, is expensive. To avoid the backfilling cost, \cite{shen2020towards} proposed Backward Compatible Training (BCT), an algorithm for training of the new model to ensure its embeddings are compatible with those of the old model. BCT succeeds in maintaining retrieval accuracy in the no-backfilling case. However, we show that BCT struggles to achieve the same performance gain as independent training for the new model (Section~\ref{sec:old-to-new-experiments}). Further, it can potentially carry unwanted biases from the old to the new model~\cite{abnar2020transferring}.\removed{This is not surprising, since training a more accurate model and compatibility with an old (less accurate) model are contradictory objectives. Further, when interoperating between the new and old model embeddings, training improvements from the new model are almost entirely absent }
\begin{figure*}[t]
    \centering
    \includegraphics[width=1.0\textwidth]{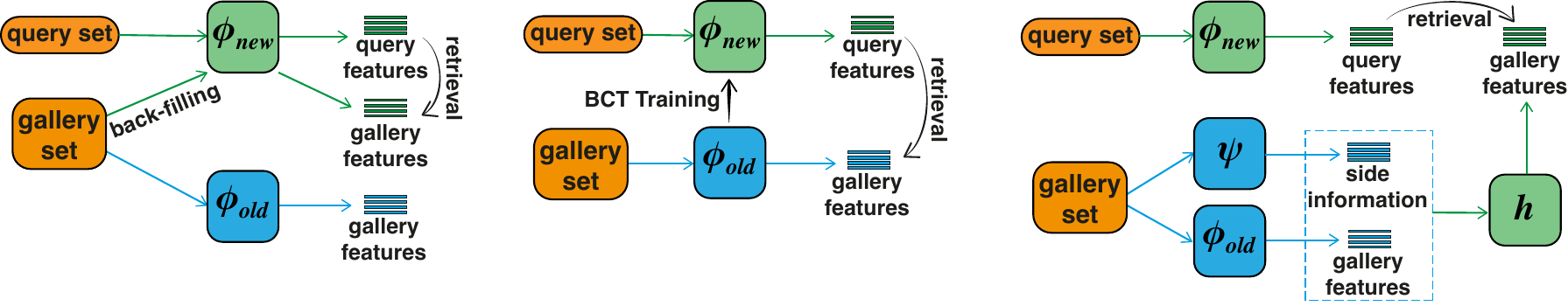}
    \caption{Independent training with back-filling \textbf{(left)}, backward compatible training \textbf{(middle)}, and forward compatible training (\fct{}) \textbf{(right)}. In \fct{}, we store side-information along with gallery features to prepare for a future embedding model.}\vspace{-10pt}
    \label{fig:methods_comparison}
\end{figure*} 

Perfect direct compatibility between new ($\phi_{new}$) and old ($\phi_{old}$) embedding models runs counter to the goal of learning better features.
Therefore, instead of trying to make $\phi_{new}$ directly compatible with $\phi_{old}$, as in \cite{shen2020towards}, we learn a transformation to map from old to new embeddings similar to~\cite{wang2020unified,meng2021learning}. 
If we are able to find a perfect transformation, the compatibility problem is solved because we can convert old to new embeddings without requiring access to the original images (no-backfilling).
However, $\phi_{old}$ possibly discards information about the image that $\phi_{new}$ does not. This information mismatch between the embeddings makes it difficult to learn a perfect transformation.


In this context, we introduce the idea of \emph{forward-compatible training} (\fct{}). This notion is borrowed from software-engineering, where software is developed with the assumption that it will be updated, and therefore is made easy to update. Here, we adopt a similar strategy for neural networks. When training the old model we know there will be an update in future, therefore we \emph{prepare} to be compatible with some unknown successor model. Future model can vary in training dataset, objective, and architecture. We propose the concept of \emph{\info{}}: auxiliary information learned at the same time as the old model which facilitates updating embeddings in the future. Intuitively, \info{} captures features of the data which are not necessarily important for the \edits{training objective of} the old model but are potentially important for the future model.

In \autoref{fig:toy} we show a toy example to demonstrate the concept. The old \edits{training objective} is to classify red objects versus blue ones. The old model therefore learns \emph{color} as the feature for each object, as this is the minimum discriminative attribute for this \edits{objective}~\cite{voita2020information}. Later, additional data is added (triangles), and the new \edits{training objective} is to classify objects based on both their shape and color. Using only what the old embeddings encode (\emph{color}), we cannot distinguish between different shapes which have the same color, e.g. blue circles and blue squares. In \fct{}, we store \info{} with the old embeddings to aid future updates. Here, \emph{shape} is the perfect \info{} to store. Note that the shape \info{} in this scenario does not help with the old \edits{training objective}, but is useful for future. Learning good \info{} is a challenging and open-ended problem since we are not aware of the future model or \edits{new training objectives}. We present our results on different possibilities for side information in Section~\ref{sec:side-info-ablate}.

To use \info{} for model compatibility, we construct a transformation function $h$, which maps from pairs of old embeddings and \info{} to new embeddings. $h$ is trained to mix the auxiliary information provided in \info{} with old embeddings to reproduce the new embedding. 
See \autoref{fig:methods_comparison} for a schematic of our setup. Using \fct{} we show significant improvements on backward-compatibility metrics (Section~\ref{sec:results}). Since $\phi_{new}$ and $\phi_{old}$ are trained independently, $\phi_{new}$ performance remains unaffected by enforced compatibility. This is in contrast to prior works \cite{shen2020towards,wang2020unified}, which have focused on making $\phi_{new}$  directly compatible with $\phi_{old}$. 

\vivek{A paragraph was deleted here, see latex document comment for original text}

\begin{figure}[t]
    \centering
    \includegraphics[width=1.0\columnwidth]{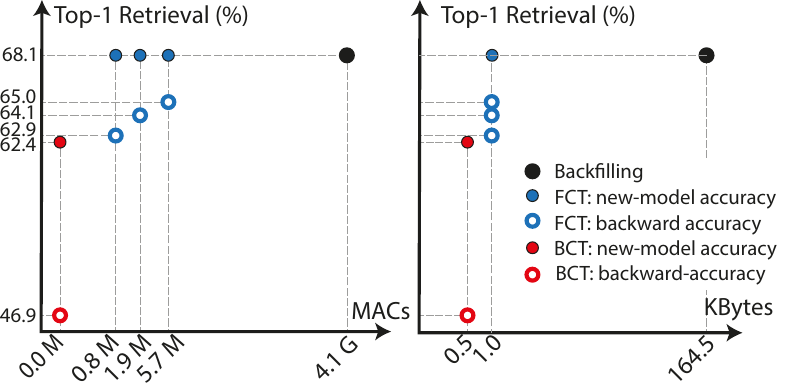}
    \caption{A comparison of compatibility methods' accuracies and associated costs. We show \textbf{(left)} Number of multiply–accumulate operations (MACs) and \textbf{(right)} data storage requirement in KBytes. Transformation models of different capacity produce different \fct{} costs and accuracies. We added 0 to the logarithmic $x$-axis \textbf{(left)} for visual comparison.}
    \label{fig:costs}
    \vspace{-5mm}
\end{figure}

We note that \fct{} requires transforming all old embeddings, which BCT avoids. This increased computational cost could be seen as a drawback of the method. However, embeddings in general have much lower dimensionality than images. In \autoref{fig:costs} we show the trade-off between computational cost (per example) and accuracy for different strategies in the ImageNet setup (with image size $3\times 224\times 224$) as described in Section~\ref{sec:results}. \fct{} does not affect new model performance, and obtains significantly higher backward accuracy for a small additional computation and storage. Note that backfilling cost (both computation and storage) scales with image resolution, but \fct{} transformation cost only depends on the embedding dimension.
\fct{} is particularly effective in the paradigm where computations take place privately on-device, trivializing its cost. We elaborate this point further in Section~\ref{sec:system}.
\vivek{cite ptflops}

Our contributions are as follows:
\begin{enumerate}[topsep=0pt,itemsep=-1ex,partopsep=1ex,parsep=1ex]
    \item We propose a new learning paradigm, forward compatible training (\fct{}), where we explicitly prepare for future model updates. \edits{Our goal is to be compatible with future models.}
    \item In the context of \fct{} for representation learning we propose \emph{\info{}}: an auxiliary features learned at the same time as $\phi_{old}$ which aids transfer to $\phi_{new}$. Intuitively, \info{} captures task-independent features of the data which are not necessarily important for the \edits{training objective} $\phi_{old}$ is trained on but are likely to be important for $\phi_{new}$.
    \item We demonstrate substantial retrieval accuracy improvement compared to BCT for various datasets: ImageNet-1k (+$18.1\%$), Places-365 (+$5.4\%$), and VGG-Face2 ($+8.3\%$). We show \fct{} outperforms the BCT paradigm by a large margin when new and old models are trained across different datasets, losses, and architectures. Unlike prior works \cite{shen2020towards,meng2021learning,wang2020unified}, models in \fct{} are trained independently, and hence their accuracies are not compromised for the purpose of compatibility.
\end{enumerate}

\section{Related Works}\label{sec:related}


\paragraph{Model Compatibility} Our closest point of comparison is \cite{shen2020towards}, which presents the problem of ``backwards-compatibility". They posit that future embeddings models should be compatible with old models when used in a retrieval setting. They present BCT, an algorithm that allows new embedding models to be compatible with old models through a joint training procedure involving a distillation loss. Other works \cite{wang2020unified,meng2021learning,chen2019r3} attempt to construct a unified representation space on which models are compatible. These procedures also modify training of individual models to ensure that they are easy to transform to this unified embedding space. This is in contrast to our work, which assumes the old and new models are trained independently.

\cite{li2015convergent} studies the relationship and degree of compatibility between two models trained on the same dataset. \cite{yosinski2014transferable} discusses the transferability of features for the same model. These are both empirical works primarily studying deep learning phenomena.

\vspace{3pt}
\noindent\textbf{Side-information} The idea of side-information is commonly discussed in the zero-shot learning literature with a different context and purpose \cite{chen2020simple}. In zero-shot learning, side-information is acquired separate from pretraining datasets, and is used to provide information about unseen classes \cite{mensink2014costa,rohrbach2010helps,elhoseiny2013write}. The recent CLIP \cite{radford2021learning} model and others \cite{elhoseiny2013write} for example use a language model as side-information to initialize their classifiers for zero-shot inference of unseen classes. Our usage of \info{} differs quite significantly, in that it is used per-example and our transfer to future tasks is not zero-shot. Further, our retrieval setup precludes the usage of side-information to define new categories.  


\vspace{3pt}
\noindent\textbf{Transfer and cross-domain learning} Transfer learning as a field is quite varied. Methods are variously classified under few-shot\cite{snell2017prototypical,li2021universal}, continual learning \cite{parisi2019continual,wortsman2020supermasks}, and life-long learning \cite{mallya2018packnet,rebuffi2017icarl}. In general, transfer learning methods seek to use knowledge learned from one domain in another to improve performance\cite{pan2009survey}. The assumption here is that transfer learning domains are related: one can exploit knowledge from one domain to aid in another domain. We have no such expectation when training old and new models. The goal of our feature transformation is not solve a new domain, but make two existing embedding models compatible, which is not a goal of transfer or cross-domain learning. \edits{In particular, even if a model can learn a new and old objective simultaneously, there is no guarantee that this model will be compatible with a model which has only learned the old objective (see Section~\ref{sec:old-to-new-experiments} for details)}. Current practice for transfer learning largely centers around various fine-tuning schemes \cite{li2020rethinking}, which makes an assumption that the architecture used in the old domain and new domain are the same. We have no constraint on old and new architectures.

\section{Method}\label{sec:method}

\subsection{Problem Setup}

A gallery set $\mathcal{G}$ is a collection of images, which are grouped into different clusters, $\{y_1\ldots y_n\}$. In visual retrieval, given a query image of a particular class, the goal is to retrieve images from $\mathcal{G}$ with the same class. A set of such query images is called the query set $\mathcal{Q}$. 

In embedding based retrieval, we have an embedding model $\phi: \mathbb{R}^{D}\to\mathbb{R}^{d}$ where $d << D$ and $D$ is the dimensionality of the input image, trained offline on some dataset $\mathcal{D}$ which embeds the query and gallery images. $\mathcal{D}$ is disjoint from $\mathcal{Q}$ and $\mathcal{G}$. Then, using some distance function $\mathfrak{D} : \mathbb{R}^d \times \mathbb{R}^d \to \mathbb{R}_{\geq 0}$, we get the closest possible image to query image $x_q\in\mathcal{Q}$ as $\argmin_{x\in\mathcal{G}} \mathfrak{D}(\phi(x_q), \phi(x))$. We use L2-distance for $\mathfrak{D}$ in our case. 

In our particular setup, we have two embedding models: the old model $\phi_{old}: \mathbb{R}^{D}\to\mathbb{R}^{d_{old}}$ and the new model $\phi_{new}: \mathbb{R}^{D}\to\mathbb{R}^{d_{new}}$, where $d_{old}, d_{new} << D$ are the embedding dimensions of old and new models, respectively. $\phi_{old}$ and $\phi_{new}$ are trained with datasets $\mathcal{D}_{old}$ and $\mathcal{D}_{new}$, respectively, using some supervised loss function.  

We assume $\phi_{old}$ is applied to every $x\in\mathcal{G}$, generating a collection of gallery embeddings, and  $\phi_{new}$ is applied to every $x_q\in\mathcal{Q}$, generating a set of query embeddings. Our goal is to design a method for performing retrieval between these two sets of embeddings without directly using the images in $\mathcal{G}$. We quantify the model compatibility performance as the accuracy of this retrieval.



\subsection{Forward Compatibility Setup}

In \fct{}, we define a function $\psi: \mathbb{R}^{D} \to \mathbb{R}^{d_{side}}$, which takes in an input from the $\mathcal{G}$ and produces our \info{}. Along with every embedding $\phi_{old}(x)$ for $x \in \mathcal{G}$ we also store the corresponding \info{} $\psi(x)$. Finally, we have a transformation model $h: \mathbb{R}^{d_{old}}\times \mathbb{R}^{d_{side}}\to\mathbb{R}^{d_{new}}$ which maps from $\phi_{old}$ and $\psi$ to $\phi_{new}$. In the \fct{} setup, we assume $\psi$ and $h$ are trained using $\mathcal{D}_{old}$ and  $\mathcal{D}_{new}$ training sets, respectively. To perform retrieval using $\psi$ and $h$, we take $\argmin_{x\in\mathcal{G}} \mathfrak{D}(h(\phi_{old}(x), \psi(x)), \phi_{new}(x))$. 

\fct{} has a few key properties and constraints:

\noindent\textbf{1.} We do not modify the training of the new model, so it gets the highest possible accuracy.

\noindent\textbf{2.} We make the old model representation compatible to that of the new model through a learned transformation.

\noindent\textbf{3.} As we discussed in the example of \autoref{fig:toy}, direct transformation from old embeddings to new may not be possible. When we train the old model, we \emph{prepare} for a future update by storing \info{} for each example in $\mathcal{G}$.

\begin{figure}[t]
\centering
    \includegraphics[width=0.7\columnwidth]{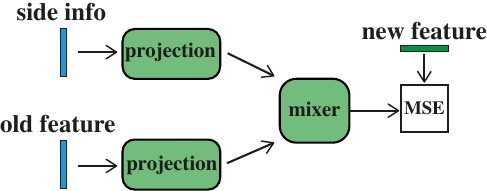}
    \caption{\fct{} transformation model architecture and training. We use an MLP with BatchNorm~\cite{ioffe2015batch} for the projection layers and the mixer. The exact layer sizes are reported in Appendix~\ref{sec:hyperparams}.}\vspace{-10pt}
    \label{fig:transfer}
\end{figure}

\subsection{Training \fct{} Transformation} \label{sec:transformation}

For each sample $x\in\mathcal{D}_{new}$, we compute the transformation $h(\phi_{old}(x), \psi(x))$. Our objective to minimize is $\ell(x) = \|h(\phi_{old}(x), \psi(x)) - \phi_{new}\|_2^2$. In our case, we write $h$ as $h_\theta$, a neural network parameterized by $\theta$. Our final optimization problem can be written as $\theta^* = \argmin_{\theta} \sum_{x\in\mathcal{D}_{new}} \ell(x)$. The transformation model consists of two projection layers corresponding to old embedding and side-information branches followed by a mixing layer to reconstruct the new embedding (\autoref{fig:transfer}). We tested several alternative objectives, including KL divergence as in \cite{hinton2015distilling}, but found L2 loss to perform the best, in contrast to \cite{shen2020towards}, which finds that using L2 loss does not allow for backwards compatibility. Recently, \cite{budnik2021asymmetric} also observed superior performance using L2 loss between features for knowledge distillation. We provide training details in the Appendix~\ref{sec:hyperparams}.




\vspace{-1pt}
\subsection{Training \fct{} Side-Information}
The ideal side-information encodes compressed features for each example with which we can reconstruct new features given the old ones. This is a challenging task since we do not know about the future task/model when learning the \info{}. The old model mainly learns features that are useful for its own training objective, and is (ideally) invariant to extraneous information. \textit{Side-information}, on the other hand, should be task independent, and capture features complementary to the old embeddings, that are useful for possible future updates.

Learning features has been extensively studied in the context of unsupervised and contrastive learning \cite{chen2020simple,bachman2019learning,gidaris2018unsupervised,noroozi2016unsupervised,zhang2016colorful,doersch2015unsupervised} when labels are not available. Here, our objective is to learn task-independent features that are not necessarily relevant for the old task, and hence the labels. This makes unsupervised learning method a particularly interesting choice to learn \info{} for \fct{}. We use SimCLR contrastive learning \cite{chen2020simple} to train $\psi$. As a constraint of our setup, we can only train $\psi$ with data available at old training. We study other choices of side-information for \fct{} in Section~\ref{sec:ablation}. We provide details of the improved SimCLR training in Appendix~\ref{sec:genclr}.

\subsection{Decentralized Design} \label{sec:system}



Compared to BCT, \fct{} comes with two additional costs: (1) when a new image is being added to the gallery-set we need to compute and store side-information, and (2) when the embedding model is updated we need to perform a transformation. (1) is a one-time computation for every new image added to the system (an incremental cost) and the side-information is small relative to the size of the image. For (2) the transformation requires access to old embeddings and side-information. This is a small computation compared to full backfilling (i.e., running the new model on all images in the gallery set) as demonstrated in \autoref{fig:costs}. 

\begin{figure}[t]
    \centering
    \includegraphics[width=1.0\columnwidth]{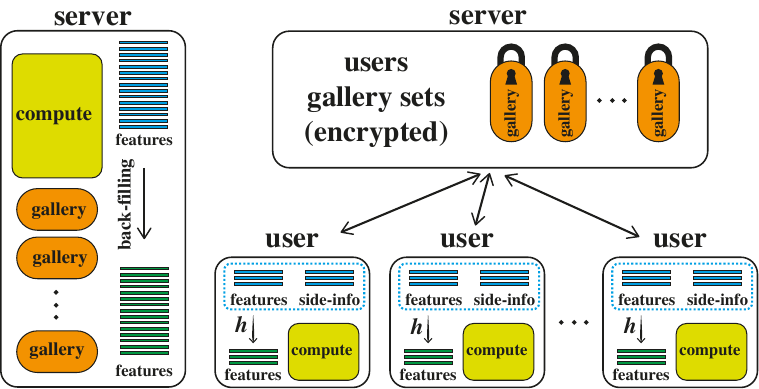}
    \caption{\textbf{(left)} A centralized system where data and compute reside on a server. Backfilling cost for such a system is mainly due to high compute. \textbf{(right)} A decentralized system where data is stored encrypted on a server, and \fct{} computations take place privately on edge devices. Storing all data on the user side is infeasible due to capacity constraints, so only side-info and embeddings are stored on device. The primary backfilling cost in this scenario is transferring images from the server to every edge device.}\vspace{-5mm}
    \label{fig:decentralized}.
\end{figure}

For many applications \fct{} can be implemented in a decentralized fashion (Figure \ref{fig:decentralized}). In this setup, we have edge devices, each with their own gallery sets.
When a new image is added, its embedding and side-information are computed and stored on device, and the raw image is encrypted and transferred to a remote server for storage.
To update the model from old to new, the \fct{} transformation runs on device. This design has three benefits: 1) The raw data always remains encrypted outside of the device without need to download it every time the model is updated. 2) Embedding and \info{} computations are privately performed on-device one-time for every new image. 3) The small \fct{} transformation computation is massively distributed on edge-devices when updating the model.
\section{Experiments}\label{sec:results}
In our experimentation, we consider different model update setups by changing training dataset, gallery/query sets, architecture, and loss. We show \fct{} consistently results in high model compatibility while not affecting accuracy of the new model.

\subsection{Evaluation Metrics}

\paragraph{Cumulative Matching Characteristics (CMC)} corresponds to top-$k$ accuracy. A similarity ordering is produced using the query embedding and every embedding in the gallery set, sorted by lowest L2 distance. If an image with the same class or identity (for face retrieval) appears in the top-$k$ retrievals, this is recorded as correct. We report CMC top-1 and 5 percentages for all models.

\vspace{3pt}
\noindent\textbf{Mean Average Precision (mAP)} is a standard metric which summarizes precision and recall metrics by taking the area under the precision-recall curve. We compute mAP@1.0, which is the average precision over recall values in [0.0, 1.0].

\vspace{3pt}
\noindent\textbf{Notation} Following \cite{shen2020towards}, we use the notation Query Embedding / Gallery Embedding to denote which model we are using for query and gallery embeddings, respectively. For example $\phi_{new} / \phi_{old}$ refers to using the new model for the query embedding and the old model for the gallery embedding. The evaluation task (retrieval from the gallery set) is fixed, while the old and new embedding models are trained differently.

All \fct{} results are \textit{backwards-compatible} with the definition provided in \cite{shen2020towards} on CMC and mAP metrics.

\subsection{Datasets}

\paragraph{ImageNet-1k}\label{sec:dat-imagenet}  \cite{imagenet} is a large-scale image recognition dataset used in the ILSVRC 2012 challenge. It has 1000 image classes. Its train set is approximately class balanced with $\sim$1.2k images per class, and its validation set is exactly class balanced with 500 images per class. We call a subset of the ImageNet-1k training set consisting of images of the first 500 classes, ImageNet-500. This subset is used for training of the old model. This is a biased split: the second 500 classes are generally more difficult (e.g. they contain all the dog breeds). For retrieval, we utilize full ImageNet-1k validation set for both the query set and gallery set. When we query with an image, we remove it from the validation set to fairly compute retrieval accuracy. 

\vspace{3pt}
\noindent\textbf{VGGFace2} \cite{cao2018vggface2} is a large-scale face recognition dataset of 3.31 million images of 9131 subjects. It is split into a train set of 8631 subjects and a validation set of 500 (disjoint) subjects. On average, there are 362.6 images per subject in the train set and exactly 500 images per subject in the validation set. The old train set is constructed from the first $10\%$ of subjects. We call this train set VGGFace2-863 and the full train set as VGGFace2-8631. For the validation, we generate a (fixed) random subsample of 50 images per subject in the validation set to perform retrieval on, as in \cite{cao2018vggface2}.

\vspace{3pt}
\noindent\textbf{Places-365} \cite{places365} is a large-scale scene recognition dataset with $\sim$1.8 million images of 365 scene categories with between 3068 and 5000 images per category. We use the first 182 classes as the old train set, which we refer to as Places-182. The validation set for Places-365 contains 36500 images, 100 images per scene category. We use the full validation set for both the query and gallery set, as with ImageNet. 



\subsection{Old to New Update Experiments}\label{sec:old-to-new-experiments}

By convention, we use the notation (Architecture)-(Embedding Dimension). We choose the embedding dimension with the highest accuracy on the new model. Our ablation on this is provided in Appendix~\ref{sec:transform-size}. Note that $(\phi_{new}/\phi_{new})$ is the upper bound and $(\phi_{old} / \phi_{old})$ is the lower bound for model compatibility. \edits{We further clarify that in each set of experiments, the evaluation criterion is the same regardless of training objective: retrieval on a fixed gallery set.}

\vspace{3pt}
\noindent\textbf{ImageNet-500 to ImageNet-1k} In this experiment, we train $\phi_{old}$ (ResNet50-128) on ImageNet-500 and $\phi_{new}$ (ResNet50-128) on ImageNet-1k using the Softmax cross-entropy loss. The side-information model $\psi$ for \fct{} is also a ResNet50-128 trained on the ImageNet-500 dataset with SimCLR~\cite{chen2020simple}. \edits{For retrieval evaluation, we use the ImageNet-1k validation set as both the query and gallery sets.} Results are provided in \autoref{tab:imagenet_result}.



\fct{} obtains substantial improvement over BCT. First, \fct{}, by construction, does not affect accuracy of the new model ($68.1\%$ top-1, which is the same as independent training). This is opposed to $5.7\%$ top-1 accuracy drop for BCT. When comparing the CMC top-1 in the compatibility setup \fct{} outperforms BCT even more significantly: $46.9\%$ to $65.0\%$, $+18.1\%$ improvement in CMC top-1. In fact, \fct{}s' top-1 compatibility accuracy ($65.0\%$) is more than the upper bound accuracy in BCT ($62.4\%$) by $2.6\%$. For reference, in the case where we use no side-info, where $\psi(x) = 0$, we get a drop of top-1 compatibility accuracy to 61.8\%, showing the importance of side-information. See Section~\ref{sec:side-info-ablate} for a more detailed ablation on the affects of side-information.

We also report accuracy of the transformed old model, $h(\phi_{old}, \psi) / h(\phi_{old}, \psi)$. We see significant improvement over the independently trained old model: $46.5\%$ to $59.3\%$. This shows that a model update with \fct{}, not only results in model compatibility, but also significantly improves quality of the old features through the transformation. This has significant consequences for practical use-cases such as embedding based clustering. We report comparisons to~\cite{meng2021learning} and \cite{wang2020unified}, modified for our setting, in Appendix~\ref{sec:lce-rbt}.

\vspace{3pt}
\noindent\textbf{Places-182 to Places-365} This experiment uses the same setup as ImageNet-500 to ImageNet-1k, except with training the old model (ResNet50-512) on Places-182 and the new model (ResNet50-512) on Places-365. The SimCLR side-information used for these experiments is trained on ImageNet-1k, showing that side-information can be transferred between domains. 
\edits{For retrieval evaluation, we use
the Places-365 validation set as both the query and
gallery sets.} These results are shown in Table~\ref{tab:places_result}. Our compatibility performance $(\phi_{new}/h(\phi_{old}, \psi))$ yields a 5.4\% improvement in CMC top-1 over BCT, even outdoing its upper bound by 0.9\% top-1.

\vspace{3pt}
\noindent\textbf{VGGFace2-863 to VGGFace2-8631} In this experiment, we train the old model (ResNet18-128) on VGGFace2-863 and the new model (ResNet50-128) on VGGFace2-8631 using the ArcFace objective~\cite{deng2019arcface}. The exact hyperparameters for these experiments is provided in Appendix~\ref{sec:hyperparams}. For retrieval evaluation, we use
the VGGFace2 validation set. Following~\cite{cao2018vggface2}, we randomly sample 50 images for each subject and use this as both the query and
gallery sets. In this experiment we use the \emph{alternate old model} side-information, another training run of the old model, differing only in SGD randomness and initialization (Section~\ref{sec:side-info-ablate}). ImageNet-1k SimCLR side-information provided marginal improvement over no side information (91.6\% to 92.0\% top-1). This makes sense because ImageNet as a domain is very distant from cropped faces. 
These results are shown in Table~\ref{tab:vggface} and are consistent with the results on other datasets.  Our compatibility performance yields a 8.3\% improvement in CMC top-1 over BCT. This shows our algorithm generalizes to models trained with objectives other than softmax cross-entropy.

\vspace{3pt}
\noindent\textbf{ImageNet-500 to ImageNet-1k w/ changing architecture}
This experiment has the same setup as ImageNet-500 to ImageNet-1k, except we consider model architecture is also changed during the update. Changing architecture when updating a model is frequent practice. We show \fct{} results in \autoref{tab:change-arch} on architectures of varying depths, structure, and embedding dimension compared to the new model. We see that as old model performance drops, compatibility performance also drops. However, in all cases compatibility performance remains quite high, still outperforming the BCT upper bound on ResNet50.
\begin{table}[t]
\centering
\resizebox{1.0\columnwidth}{!}{
\begin{tabular}{lccc}
  \toprule
Method & Case &
\begin{tabular}{c}
     CMC  \\
top-1|top-5 \%
\end{tabular}
& mAP@1.0 \\
\toprule
\multirow{3}{*}{Independent} & $\phi_{old} / \phi_{old}$ & 46.5 | 64.6  & 29.9\\
& $\phi_{new} / \phi_{old}$ & 0.1 | 0.5 & 0.003\\
& $\phi_{new} / \phi_{new}$ & 68.1 | 84.4 & 45.0\\
\midrule
\multirow{2}{*}{BCT\cite{shen2020towards}} & $\phi^{BCT}_{new} / \phi_{old}$ & 46.9 | 65.4 & 30.1\\
& $\phi^{BCT}_{new} / \phi^{BCT}_{new}$ & 62.4 | 81.9 & 41.1\\
\midrule
\multirow{3}{*}{\fct{} (ours)} 
& $h(\phi_{old}, \psi) / h(\phi_{old}, \psi)$ & {\bf 59.3 | 76.4} & {\bf 41.3}\\
& $\phi_{new} / h(\phi_{old}, \psi)$ & {\bf 65.0 | 82.3} & {\bf 43.6}\\
& $\phi_{new} / \phi_{new}$ & \textbf{68.1 | 84.4} & \textbf{45.0}\\
\bottomrule
  \end{tabular}}
\caption{Comparison of different compatible training methods. The old model is trained on the ImageNet-500, and new model on the ImageNet-1k. The ImageNet validation set is used as gallery and query sets. Both models are ResNet50-128.}\vspace{-10pt}
  \label{tab:imagenet_result}
\end{table}

\begin{table}[t]
\centering
\resizebox{1.0\columnwidth}{!}{
\begin{tabular}{lccc}
  \toprule
Method & Case &
\begin{tabular}{c}
     CMC  \\
top-1|top-5 \%
\end{tabular}
 & mAP@1.0 \\
\toprule
\multirow{3}{*}{Independent} & $\phi_{old} / \phi_{old}$ & 29.6 | 58.2 & 11.6 \\
& $\phi_{new} / \phi_{old}$ & 0.3 | 1.5 & 0.12\\
& $\phi_{new} / \phi_{new}$ & 37.0 | 65.1 & 17.0\\
\midrule
\multirow{2}{*}{BCT\cite{shen2020towards}} & $\phi^{BCT}_{new} / \phi_{old}$ & 30.4 | 58.7 & 12.6\\
& $\phi^{BCT}_{new} / \phi^{BCT}_{new}$ & 34.9 | 64.2 & 16.0\\
\midrule
\multirow{3}{*}{\fct{} (ours)} 
& $h(\phi_{old}, \psi) / h(\phi_{old}, \psi)$ & {\bf 34.0 | 62.3} & {\bf 17.3}\\
& $\phi_{new} / h(\phi_{old}, \psi)$ & {\bf 35.8 | 64.5} & {\bf 18.1}\\
& $\phi_{new} / \phi_{new}$ & \textbf{37.0 | 65.1} & \textbf{17.0}\\
\bottomrule
  \end{tabular}}
\caption{Comparison of different compatible training methods. The old model is trained on Places-182, and new model on Places-365. The Places-365 validation set is used as gallery and query sets. Both models are ResNet50-512.}
  \label{tab:places_result}
\end{table}

\begin{table}[t]
\centering
\resizebox{1.0\columnwidth}{!}{
\begin{tabular}{lccc}
  \toprule
Method & Case &
\begin{tabular}{c}
     CMC  \\
top-1|top-5 \%
\end{tabular}
 & mAP@1.0 \\
  \toprule
\multirow{3}{*}{Independent} & $\phi_{old} / \phi_{old}$ & 84.0 | 93.5 & 44.6 \\
& $\phi_{new} / \phi_{old}$ & 0.2 | 1.0 & 0.2\\
& $\phi_{new} / \phi_{new}$ & 96.6 | 98.4 & 77.8\\
\midrule
\multirow{2}{*}{BCT\cite{shen2020towards}} & $\phi^{BCT}_{new} / \phi_{old}$ & 84.2 | 93.9 & 45.0\\
& $\phi^{BCT}_{new} / \phi^{BCT}_{new}$ & 95.1 | 98.3 & 74.3\\
\midrule
\multirow{3}{*}{\fct{} (ours)} 
& $h(\phi_{old}, \psi) / h(\phi_{old}, \psi)$ & {\bf 87.2 | 94.9} & {\bf 53.4}\\
& $\phi_{new} / h(\phi_{old}, \psi)$ & {\bf 92.5 | 97.5} & {\bf 62.6}\\
& $\phi_{new} / \phi_{new}$ & \textbf{96.6 | 98.4} & \textbf{77.8}\\
\bottomrule
  \end{tabular}}
\caption{Comparison of different compatible training methods on VGGFace2. The old model is trained on VGGFace2-863, and the new model is trained on VGGFace2-8631 with the ArcFace training loss. The architecture is ResNet18-128 for the old model and ResNet50-128 for the new model.}
  \label{tab:vggface}
\end{table}

\begin{table}[t]
\centering
\resizebox{1.0\columnwidth}{!}{
\begin{tabular}{lccc}
  \toprule
Old Architecture & Case &
\begin{tabular}{c}
     CMC  \\
top-1|top-5 \%
\end{tabular}
 & mAP@1.0 \\
  \toprule
N/A & $\phi_{new} / \phi_{new}$ & 68.1|84.4 & 45.0\\
\midrule
\multirow{3}{*}{ResNet18-128~\cite{he2016deep}} & $\phi_{old} / \phi_{old}$ & 40.1|60.1 & 19.9 \\
& $h(\phi_{old}, \psi) / h(\phi_{old}, \psi)$ & 57.0|75.0 & 39.4 \\
& $\phi_{new} / h(\phi_{old}, \psi)$ & 64.2|82.1 & 42.6\\
\midrule
\multirow{3}{*}{MobileNet-128~\cite{howard2017mobilenets}} & $\phi_{old} / \phi_{old}$ & 42.3|62.7 & 20.3 \\
& $h(\phi_{old}, \psi) / h(\phi_{old}, \psi)$ & 57.6|75.7 & 39.8 \\
& $\phi_{new} / h(\phi_{old}, \psi)$ & 64.3|82.1 & 42.9\\
\midrule
\multirow{3}{*}{ResNet50-64~\cite{he2016deep}} & $\phi_{old} / \phi_{old}$ & 44.7|62.7 & 30.6 \\
& $h(\phi_{old}, \psi) / h(\phi_{old}, \psi)$ & 56.6|74.1 & 39.0 \\
& $\phi_{new} / h(\phi_{old}, \psi)$ & 63.6|81.9 & 42.0\\
\bottomrule
  \end{tabular}}
\caption{\fct{} compatibility results when old and new architectures are different. For all cases, the new model is ResNet50-128. The old model is trained on the ImageNet-500, and new model on the ImageNet-1k. The ImageNet validation set is used as gallery and query sets.}\vspace{-10pt}
  \label{tab:change-arch}
\end{table}

\vspace{3pt}
\noindent\textbf{ImageNet-500 to ImageNet-1k evaluated on Places-365} In this experiment we evaluate the same embedding models as in the ImageNet-500 to ImageNet-1k experiment using Places-365 validation set as gallery and query sets. This is a challenging case since Places-365 is a different domain from ImageNet. This is clear from the drop in old and new model retrieval performance from ImageNet to Places365 in this setting: 46.3\% to 14.5\% and 68.1\% to 21.9\% top-1, respectively. With \fct{} we obtain 20.3\% top-1 compatibility performance, $\phi_{new} / h(\phi_{old}, \psi)$, only 1.6\% below the accuracy of the new model. This shows generalization of the side-information and transformation models to out-of-domain data. In Appendix~\ref{sec:ood}, we report more results with this setup. We show that when the training objectives of $\phi_{old}$ and $\phi_{new}$ are disjoint (trained on disjoint training sets), a particularly difficult scenario, we can maintain model compatibility with performance very close to that of $\phi_{new}$.


\subsection{Sequence of Model Updates}\label{sec:sequence}
Consider a sequence of model updates: $\text{v}_{1}\rightarrow \text{v}_{2}\rightarrow \ldots \rightarrow \text{v}_{n} $. The first model, v$_1$, directly computes features and side-information from the gallery set. The v$_i$ embedding and side-information models, denoted by $\phi_{i}$ and $\psi_{i}$, respectively, are trained independently on the v$_i$ training set.
When updating from v$_{i}$ to v$_{i+1}$, we apply \fct{} transformation on both features and side-information of v$_{i}$ to make them compatible with those of v$_{i+1}$:
\begin{equation*}
    \phi_{i+1} \xleftrightarrow{\text{compatible}} h_{i+1}(\phi_{i}, \psi_{i}),~
    \psi_{i+1} \xleftrightarrow{\text{compatible}} g_{i+1}(\phi_{i}, \psi_{i})
\end{equation*}
We use the same architectures for $h_{i+1}$ and $g_{i+1}$ transformations, as shown in \autoref{fig:transfer}. Both transformations are trained using the MSE loss over the v$_{i+1}$ training set, the same process as in Section~\ref{sec:transformation}. 

\begin{figure}[t]
    \centering
    \includegraphics[width=1.0\columnwidth]{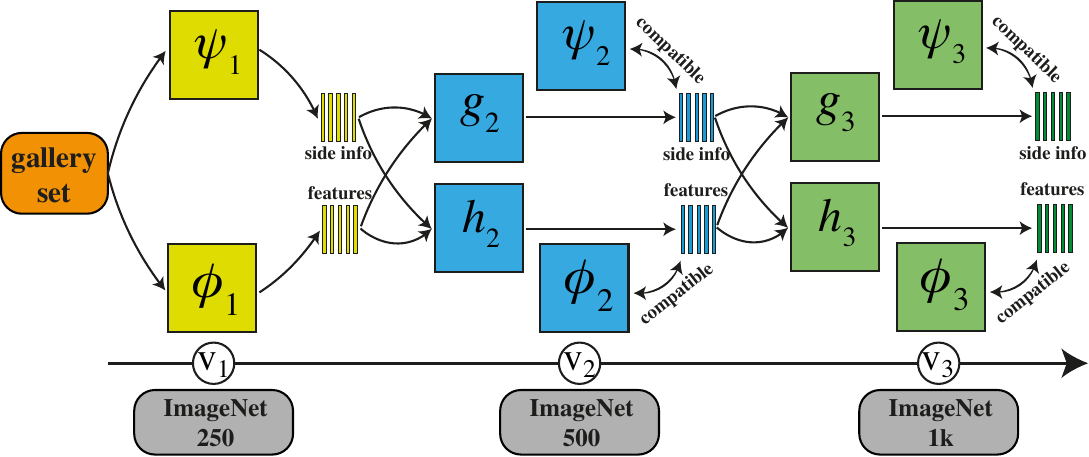}
    \caption{\fct{} for a sequence of model updates. Here, we transform both side-information and embedding to be compatible with their next versions.}\vspace{-10pt}
    \label{fig:sequence}
\end{figure}

Here, we experiment with a sequence of three model updates: the v$_1$ model is trained on the ImageNet-250 training set (a quarter of the full ImageNet training set). v$_2$ and v$_3$ models have the same setup as v$_1$ but with larger training sets: ImageNet-500 and ImageNet-1k, respectively. The v$_i$ side-information model is a ResNet50-128 backbone trained on v$_i$ training set using SimCLR. We compare two \fct{} scenarios: (i) v$_1$ is first updated to v$_2$ and then v$_2$ is updated to v$_3$ as shown in \autoref{fig:sequence}, and (ii) v$_1$ is directly updated to v$_3$ using $h_{1\to3}$ transformation.

We show compatibility results in \autoref{tab:sequence}. Last two rows correspond to scenario (i) and (ii), respectively. \fct{} demonstrates a great performance ($57.4\%$ top-1) even in the two-hop update, Scenario (i). Scenario (ii) is slightly more accurate ($+1.5\%$), but in practice requires extra storage of v$_1$'s feature and side-information (even after updating to v$_i$ with $i>1$). In Scenario (i) all v$_i$ features and side-information are replaced when updating to v$_{i+1}$ as shown in \autoref{fig:sequence}.

If we do not store side-information, we observe significant drop in compatibility performance for both scenarios: \emph{57.4\%} to \emph{44.9\%} in Scenario (i) and \emph{59.9\%} to \emph{53.9\%} in Scenario (ii). This shows storing side-information is crucial for compatibility performance in a sequence of model updates. In Appendix~\ref{sec:long}, we further demonstrate the importance of side-information by scaling to a longer sequence of small updates using subsets of the CIFAR-100 dataset\cite{krizhevsky2009learning}.

\begin{table}[t]
\centering

\resizebox{1.0\columnwidth}{!}{\begin{tabular}{ccc}
  \toprule
Case & 
CMC top-1 | top-5 \%
&
mAP@1.0
\\
  \toprule
 $\phi_{1} / \phi_{1}$ &  29.6 | 44.1 & 15.5\\
 $\phi_{2} / \phi_{2}$ &  46.5 | 65.1 & 28.7\\
 $\phi_{3} / \phi_{3}$ &  68.1 | 84.4 & 45.0\\
 $ \phi_{2} / h_2(\phi_{1}, \psi_{1})$ &  40.5 | 58.8 & 26.4\\
 $\phi_{3} / h_3(\phi_{2}, \psi_{2})$ &  65.0 | 82.3 & 43.6\\
  $\phi_{3} /  h_3(
  h_2(\phi_{1}, \psi_{1}),
  g_2(\phi_{1}, \psi_{1})
  )$ &  57.4 | 78.0 & 33.5\\
 $\phi_{3} / h_{1\to3}(\phi_{1}, \psi_{1})$ &  59.9 | 79.3 & 37.0\\
 
\bottomrule
  \end{tabular}}
\caption{Model compatibility in sequence of updates. We see only a small degradation in performance in performing a sequence of updates v$_1\to$ v$_2\to$ v$_3$ versus going directly from v$_1\to$ v$_3$, corresponding to the last two rows.}\vspace{-10pt}
  \label{tab:sequence}
\end{table}

\section{Analysis}\label{sec:analysis}\label{sec:ablation}
\subsection{Types of Side-Information}\label{sec:side-info-ablate}


\noindent\textbf{No side-information} We use no side-information as a simple baseline. In terms of implementation, we pass the zero-vector into the projection layer in \autoref{fig:transfer}. 


\noindent\textbf{Autoencoder} We train a simple autoencoder with L2 reconstruction loss. The encoder and decoder architectures are convolutional, and based on MobileNetv2~\cite{sandler2018mobilenetv2}. The exact architectures are reported in Appendix~\ref{sec:hyperparams}. 

\noindent\textbf{Alternate Old Model} \removed{This is the simplest form of side-information that we consider.} With this method, we train another version of $\phi_{old}$, with the only difference being randomness from data order and model initialization. This is very similar to ensembling~\cite{fort2019deep}. The intuition is that each model captures a different facet of the data. We denote this model as $\phi_{old}^{alt}$.

\noindent\textbf{Alternate Model + Mixup} We perform the same training as for $\phi_{old}$ but with Mixup augmentation~\cite{zhang2017mixup} applied only on images (for labels we use one-hot vectors). This encourages learning features of the data different from the old model, which will aid in transformation as they capture different invariances of the data.

\noindent\textbf{Contrastive Model} Here we train a SimCLR \cite{chen2020simple} self-supervised contrastive learning model to use as side-information. Taking the previous ideas to their natural extreme, a self-supervised contrastive approach directly captures invariances in the data, which will be useful for transfer even if it is not as useful for retrieval. 




\begin{table}[t]
\centering

\resizebox{1.0\columnwidth}{!}{\begin{tabular}{lccccc}
  \toprule
Side-Information $(\psi)$ &
\begin{tabular}{c}
     $\phi_{new} / h(\phi_{old}, \psi)$  \\
CMC top-1 | top-5 \%
\end{tabular}
& mAP@1.0
& CKA\\
  \toprule
None &  61.8 | 80.5 & 39.9  & 0.0\\
\midrule
Autoencoder &  62.0 | 80.9 & 40.5 & 0.04\\
$\phi_{old}^{alt}$ & 63.5 | 81.6 & 42.1 & 0.70\\
$\phi_{old}$ w/ Mixup & 63.8 | 81.9 & 42.4 & 0.66\\
SimCLR-128-ImageNet-500 & 65.0 | 82.3 & 43.6 & 0.27\\
\midrule
SimCLR-128-ImageNet-1k & 66.5 | 83.3 &46.6 & 0.25\\
\bottomrule
  \end{tabular}}
\caption{Comparison of different kinds of side-info. Both old and new architectures are ResNet-50 with feature size 128. The old model is trained on ImageNet-500 and the new model is trained on ImageNet-1k}\vspace{-10pt}
  \label{tab:side-info-ablation}
\end{table}


\subsection{Side-information Ablation Results}
Results are shown in \autoref{tab:side-info-ablation}. First note that no side-information results in 61.8\% top-1 retrieval accuracy, which is 14.9\% greater than BCT in this setting. This demonstrates the strength of our transformation setup even without side-information. The autoencoder provides a slight improvement over no side-information. This make sense considering much of the information of the autoencoder is encoding is contextually dependent on its decoder, and is therefore hard to extract \textit{a priori}. 

The $\phi_{old}^{alt}$ and the $\phi_{old}^{alt}$ with Mixup both provide substantial improvements over no side-information. It is well-known in deep learning literature that different runs of SGD for the same model produce diverse predictions \cite{mania2019model, fort2019deep}, which is particularly useful for ensembling. Here we show that we can leverage this fact for side-information. It's important to note that simply applying a transformation from ResNet50-256 with no side-info trained on ImageNet-500 to the new model only yields 62.0\% CMC top-1, which is less than the improvement we get from training a second old model. Mixup regularization gets a further 0.3\% improvement over $\phi_{old}^{alt}$, even though it does not provide a better model for retrieval, with 46.4\% $(\phi_{old} / \phi_{old})$ CMC top-1.

The SimCLR model with embedding dimension of $128$ trained on half of the ImageNet training set, denoted by SimCLR-128-ImageNet-500, outperforms other side-information choices considered (+1.2\% over Mixup). This is despite the fact that CMC top-1 with just the SimCLR model is only 35.41\%, far worse than even $\phi_{old}$ at 46.5\%. What's even more interesting is that when we train our transfer model with only SimCLR side-information and no $\phi_{old}$ we get 63.2\% top-1 retrieval accuracy, better than no side-info. This demonstrates the efficacy of contrastive features for forward transformation. Even given the large gap in retrieval performance between $\phi_{old}$ and SimCLR, transfer performance for SimCLR is still superior, validating our intuition that it is important for side-information to capture all the invariances of the target data source. 


We also consider a case that for the old model training we have access to the next version training set (ImageNet-1k), yet without labels. We use the unlabeled dataset to train the side-information model, shown by SimCLR-128-ImageNet-1k. This side-information results in $66.5\%$ backward compatibility accuracy, $+1.5\%$ boost compared to the case when SimCLR-128-ImageNet-500 is used as the side-information model.

\subsection{Centered Kernel Alignment (CKA) Analysis}
CKA \cite{kornblith2019similarity} is a similarity index identifying correspondences between representations. We show CKA with linear kernel between the old model features and side-information over the gallery set (the ImageNet validation set) in \autoref{tab:side-info-ablation}. We use CKA as a measure of complementarity of the side-information to the old model. Evidently, using mixing to train an alternate old model results in lower CKA (0.7 to 0.66), and hence higher compatibility accuracy when used as a side-information. More interestingly, the SimCLR side-information model has significantly less CKA with the old model (0.27). This demonstrates features learned by SimCLR are \emph{complementary} to those of the old model. Hence, combining them in \fct{} transformation results in excellent accuracy (65.0). Note that no side-information (the zero vector) and the Autoencoder have very low CKA but poor performance, demonstrating that this is not the only factor important to selecting a good model for side-information. \vspace{-5pt}
\section{Conclusion}\label{sec:conclusion}

In this paper we study the compatibility problem in representation learning. We presented a new learning paradigm: forward compatible training, where we prepare for the next version of the model when training the current model. \fct{} enables efficient embedding updates to ensure compatibility with future models in large retrieval systems. We have shown that it is important to learn side-information to transform old features to new embedding models. Side-information captures features of the data which might not be useful for the old task, but are potentially important in the future. We demonstrate that contrastive learning methods work well to train side-information models. Through experimentation across multiple datasets, architectures, and losses, we share insights into necessary components important to the design of model-to-model transformation using side-information. Finally, by ensuring that training of the old and new embedding models are completely independent, we do not degrade model performance or retain biases from old model training, a problem in prior compatibility literature.

\section{Limitations}

\fct{} provides a mechanism to make old embeddings compatible with their future version. As a result, after transforming gallery embeddings, only downstream models which are \textit{non-parametric} will not need to be updated (e.g. nearest neighbor retrieval in our setup). This is not a requirement with BCT since new embeddings are directly comparable with old embeddings. However, the convenience of not updating the downstream task model comes with limited accuracy improvement when updating the embedding model. We have not evaluated on cross-domain compatibility: when the old and new model are trained on widely different domains. In practice, it is unclear how likely this scenario is, but we suspect it will make learning a transformation and useful side-information more difficult.


\section*{Acknowledgements}

We would like to thank Floris Chabert and Vinay Sharma for their advice and discussions, and Jason Ramapuram and Dan Busbridge for their help with self-supervised trainings.


{\small
\bibliographystyle{ieee_fullname}
\bibliography{egbib}
}

\clearpage
\appendix

\begin{table*}[t]
\centering
\resizebox{1.8\columnwidth}{!}{
\begin{tabular}{@{}ccccccc@{}}
\toprule
     \multirow{2}{*}{$\mathcal{D}_{old}$} &   \multirow{2}{*}{$\mathcal{D}_{new}$} & \multirow{2}{*}{Embedding Dimension}                     & \multicolumn{2}{c}{$\phi_{new} / h(\phi_{old})$}             & \multicolumn{2}{c}{$h(\phi_{old}) / h(\phi_{old})$}          \\\cmidrule(l){4-7} 
 &  &  & \multicolumn{1}{c}{CMC top 1|5 \%} & \multicolumn{1}{c}{mAP \%} & \multicolumn{1}{c}{CMC top 1|5 \%} & \multicolumn{1}{c}{mAP \%} \\ \midrule
ImageNet-500 & ImageNet-1k &   128 &  61.8 | 80.5 & 39.9 & 51.9 | 69.8 & 36.1 \\
ImageNet-500 & ImageNet-1k &   256 &  62.0 | 80.9 & 38.7 & 53.8 | 71.9 & 35.8 \\
Places-182   & Places-365 & 512 & 34.8 | 63.6 & 17.5 & 31.9 | 60.4 & 16.4\\
VGGFace2-863 & VGGFace2-8631 & 128 &  91.5 | 97.3 & 60.1 & 84.0 | 93.4 & 47.7 \\\bottomrule
\end{tabular}}
\caption{Detailed results in the ``no side-information" case. We see a substantial improvement in retrieval performance across all datasets.}
  \label{tab:no_side_info}
\end{table*}

Our appendix is organized as follows:

\begin{enumerate}
    \item Appendix~\ref{sec:no-side-info} contains detailed experimental results in the ``no-side information" case. We show that adding side-information outperforms no side-information in all cases, including when the embedding for ``no-side information" has dimension equal to that of (side-information + embedding).
    \item Appendix~\ref{sec:ood} shows performance of transformation with side-information on out-of-distribution datasets.
    \item Appendix~\ref{sec:degradation} shows a more detailed analysis of performance degradation of the Backwards Compatible Training (BCT) algorithm when applied to $\phi_{new}$ on ImageNet.
    \item Appendix~\ref{sec:hyperparams} reports hyperparameters and training details across all experiments.
    \item Appendix~\ref{sec:transform-size} reports the results of our ablation for the transformation function on both capacity and loss function. 
    \item Appendix~\ref{sec:embed-dim} reports details about embedding dimension ablations for $\phi_{new}$ and $\phi_{old}$ across datasets.
    \item Appendix~\ref{sec:genclr} shows our results with highly compressed side-information.
    \item \edits{Appendix~\ref{sec:long} shows our results on using \fct{} with a long sequence of small updates.}
    \item \edits{Appendix~\ref{sec:disjoint} shows our results on using \fct{} between models with disjoint training objectives.}
    \item \edits{Appendix~\ref{sec:lce-rbt} we report comparisons to~\cite{meng2021learning} and \cite{wang2020unified}, modified for our setting.}
    \item Appendix~\ref{sec:resources} reports the licenses of all the resources used.
\end{enumerate}

\section{Comparison with No Side-Information}\label{sec:no-side-info}

\subsection{Standard Setup}

\autoref{tab:no_side_info} contains detailed results for ``no side-information" in the standard setup. This corresponds to Tables~\ref{tab:imagenet_result},~\ref{tab:places_result}, and~\ref{tab:vggface}. In terms of implementation, we use $\psi(x) = 0$ as for our side-information function. We see that side-information improves model performance across all datasets. In particular, we see a CMC top-1 improvement of +3.2\% on ImageNet, +1.0\% on Places, and +1.0\% on VGGFace2. The magnitude of improvement on $h(\phi_{old} /\psi) / h(\phi_{old}, \psi)$, a measure of how much our transformation improves the old features, is even greater, with CMC top-1 improvements of +7.4\% on ImageNet, +2.1\% on Places, and +3.2\% on VGGFace2. 

Finally, an argument could be made that the overall feature dimension is simply larger, when we consider side-information feature size together with the embedding dimension feature size. To show the benefit of our side-information over simply increasing the embedding dimension size, we show that a 256-dimensional feature vector for ImageNet only results in a 0.2\% improvement in CMC top-1, far from the improvement which comes from having good side-information (+3.2\%).

\subsection{Sequence of Model Updates}

\begin{table}[t]
\centering

\resizebox{1.0\columnwidth}{!}{\begin{tabular}{ccc}
  \toprule
Case & 
CMC top-1 | top-5 \%
&
mAP@1.0
\\
  \midrule
 $\phi_{1} / \phi_{1}$ &  29.6 | 44.1 & 15.5\\
 $\phi_{2} / \phi_{2}$ &  46.5 | 65.1 & 28.7\\
 $\phi_{3} / \phi_{3}$ &  68.1 | 84.4 & 45.0\\\midrule
 $ \phi_{2} / h_2(\phi_{1})$ &  36.5 | 53.5 & 23.7\\
 $\phi_{3} / h_3(\phi_{2})$ &  61.8 | 80.5 & 39.9\\\midrule
  $\phi_{3} /  h_3(
  h_2(\phi_{1}),
  g_2(\phi_{1})
  )$ &  44.9 | 68.0 & 26.4 \\
 $\phi_{3} / h_{1\to3}(\phi_{1})$ &  53.9 | 74.2 & 30.6\\
 
\bottomrule
  \end{tabular}}
\caption{Model compatibility in sequence of updates with no side-information (see \autoref{tab:sequence} for results with side-information). We see a large performance degradation v$_1\to$ v$_3$ versus with side-information (-6.0\% CMC top-1) and an even larger degradation in the sequential case v$_1\to$ v$_2\to$ v$_3$ (-12.5\% CMC top-1), showing that side-information is crucial for sequential updates to prevent feature drift.}\vspace{-10pt}
  \label{tab:sequence_no_side}
\end{table}

\autoref{tab:sequence_no_side} contains results corresponding to the sequence of model updates case (see Section~\ref{sec:sequence}) with no side-information. As in the previous section, this is implemented by setting $\psi(x) = 0$ for all inputs $x$. We denote this case $h(\phi)$ instead of $h(\phi, \psi)$ for embedding model $\phi$ and transformation model $h$. All other notation is consistent with Section~\ref{sec:sequence}. 

\section{Out-of-distribution Side-information performance}\label{sec:ood}

\begin{table}[H]
\centering

\resizebox{0.9\columnwidth}{!}{\begin{tabular}{lcccc}
  \toprule
Side-Info &
\begin{tabular}{c}
     $(\phi_{new} / h(\phi_{old}, \psi))$  \\
CMC top-1 | top-5 \%
\end{tabular}
& mAP@1.0\\
  \midrule
None ($\phi_{old}$) & 14.5 | 35.2 & 3.7\\\midrule
None &  16.9 | 40.7 & 5.6 \\
Autoencoder & 17.4 | 41.6 & 5.8\\
$\phi_{old}^{alt}$ & 18.4 | 43.1 & 6.3\\
SimCLR & 20.3 | 45.5 & 7.1\\\midrule
None ($\phi_{new}$) & 21.9 | 46.8 & 7.1\\
\bottomrule
  \end{tabular}}
\caption{Using the same ImageNet embedding models $\phi_{old}$ and $\phi_{new}$ we analyze the same side-information strategies as \autoref{tab:side-info-ablation} but instead evaluating on out of distribution data: the Places-365 validation set. We show that the same trends hold even out of distribution. Note that the retrieval performance is quite poor since ImageNet and Places-365 are very different domains.\vivek{candidate to move to supplementary}}
  \label{tab:places-side-info-ablation}
\end{table}

Here we provide a more detailed ablation of the out-of-distribution retrieval performance for various side-information transformation methods originally presented in Section~\ref{sec:old-to-new-experiments}. We present these results in \autoref{tab:places-side-info-ablation}. We see the same trends from \autoref{tab:side-info-ablation} repeated on this very different domain.

\section{BCT New Model Biased Performance Degradation}\label{sec:degradation}

\begin{figure}[t]
    \centering
    \includegraphics[width=\columnwidth]{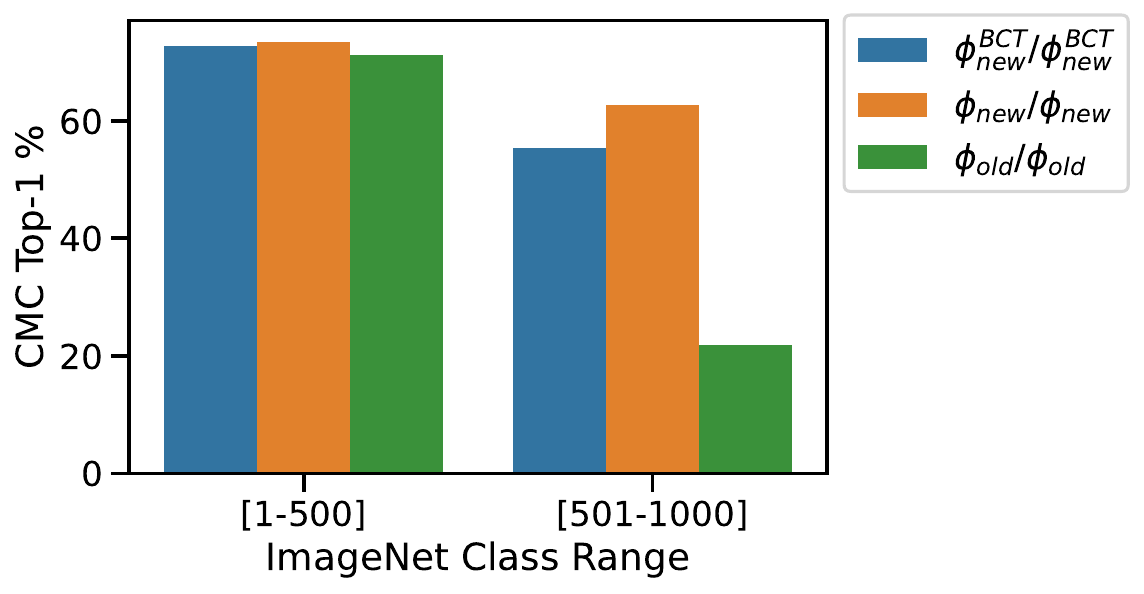}
    \caption{CMC Top-1 performance on the validation set of ImageNet for various cases by class range. Note that $\phi_{old}$ has not seen any data from classes [501-1000]. Note how $\phi_{new}^{BCT}$ performance is degraded primarily for the classes on which $\phi_{old}$ also performs poorly. This shows that the old model biases training of the new model for BCT.}
    \label{fig:degradation}\vspace{-10pt}
\end{figure}

In \autoref{fig:degradation} we show how BCT training of the new model biases it towards the old model. Since the old model has not seen any data from ImageNet classes [501-1000], it naturally performs quite poorly on these classes. However, performance on classes [1-500] is quite similar across $\phi_{new}$ and $\phi_{old}$. $\phi_{new}^{BCT}$ sees some improvement on classes [501-1000], however it performs 6.4\% worse than $\phi_{new}$ on these classes. Given that its accuracy is similar to that of $\phi_{old}$ and $\phi_{new}$ on classes [1-500], we can conclude that the effect of BCT training is to bias $\phi_{new}^{BCT}$ to perform more similarly to $\phi_{old}$, i.e. poorly on [501-1000]. Transferring unwanted biases from old to new model is a crucial drawback of the BCT method.

\section{Architectures, Hyperparameters, and Training}\label{sec:hyperparams}

\subsection{Architecture Details}

\noindent\textbf{Embedding Models} For our embedding model architectures, we use the standard ResNet~\cite{he2016deep} and MobileNetV1~\cite{howard2017mobilenets} with penultimate layers modified to output 128 dimension embeddings (for VGGFace2 and ImageNet) and 512 dimension embeddings (for Places-365). We find that this architecture modification performs either the same or better than their higher dimension counterparts. We also observed that directly modifying the output dimension of the penultimate layer performs equivalently to projecting the output of the penultimate layer to a lower dimension, as in~\cite{shen2020towards}. 

\vspace{3pt}
\noindent\textbf{Transformation Models} See Figure~\ref{fig:transfer} for the basic transformation structure. The projection units in this diagram are MLPs with architecture: $\text{Linear}(d \to 256) \to \text{BatchNorm} \to \text{ReLU} \to \text{Linear}(256\to 256) \to \text{BatchNorm}\to\text{ReLU}$, where $d$ is the dimension for either the embedding ($d_{old}$) or side-information ($d_{side}$). The linear layers use bias, however preliminary experiments have shown this is not necessary. The outputs of both projection branches are concatenated together before being passed to the ``Mixer", which has architecture $\text{Linear}(512\to 2048) \to \text{BatchNorm} \to \text{ReLU} \to \text{Linear}(2048\to 2048) \to \text{BatchNorm} \to \text{ReLU} \to \text{Linear}(2048\to d_{new})$.
\subsection{Model Selection Details}

We did not perform hyperparameter tuning on transformation architecture training. We reused these ImageNet hyperparameters on the other datasets and embedding dimension sizes. The hyperparameters for ResNet and MobileNet training were taken from ResNetV1.5~\cite{repo}, an improved training setup for ResNet. Hyperparameter details for this setup are provided in the in the individual dataset sections. 

\subsection{Hardware Details}

We trained all of our models on 8 Nvidia V100 GPUs with batch size 1024. We have found that it is possible to decrease the batch size proportionally with the learning rate to use with fewer resources (e.g. batch size 1024 with learning rate 1.024 corresponds to batch size 256 with learning rate 0.256) without a drop in performance. 
\begin{table*}[t]
\centering
\resizebox{1.6\columnwidth}{!}{\begin{tabular}{@{}ccccccc@{}}
\toprule
\multirow{2}{*}{Embedding Size} & \multicolumn{2}{c}{ImageNet-\{500, 1k\}} & \multicolumn{2}{c}{Places-\{182, 365\}} & \multicolumn{2}{c}{VGGFace2-\{836, 8631\}} \\ \cmidrule(l){2-7} 
               & $\phi_{old} / \phi_{old}$         & $\phi_{new} / \phi_{new}$        & $\phi_{old} / \phi_{old}$       & $\phi_{new} / \phi_{new}$       & $\phi_{old} / \phi_{old}$          & $\phi_{new} / \phi_{new}$         \\ \midrule
128 & 46.5 & \textbf{68.0} & 29.5 & 21.9 & 84.0 & \textbf{96.6}\\
256 & 48.0 & 67.8 & 29.7 & 23.2 & 83.8 & 95.7 \\
512 & 49.1 & 67.4 & 29.6 & \textbf{37.0} & 83.6 & 96.4\\
1024 & 49.0 & 67.1 & 29.5 & 36.7 & 84.1 & 95.9 \\
2048 & 48.5 & 67.9 & 29.8 & 37.0 & 83.9 & 95.1 \\ \bottomrule
\end{tabular}}
\caption{Embedding dimension size ablation on ImageNet, Places-365, and VGGFace2 for CMC top-1. The $\phi_{old}$ architectures are ResNet50 for Places and ImageNet and ResNet18 for VGGFace2. The $\phi_{new}$ architectures are ResNet50 for all datasets. We train $\phi_{old}$ on ImageNet-500, Places-182, and VGGFace2-863 and $\phi_{new}$ on ImageNet-1k, Places-365, and VGGFace2-8631, respectively. We chose to use the dimension corresponding to the best performing $\phi_{new}$ for each dataset (bolded). In the case of a tie, we chose the lower dimensional model.}\label{tab:emb-dim-ablation}
\end{table*}
\subsection{Side-Information}

\subsubsection{SimCLR}

\noindent\textbf{Architecture} We use a standard ResNet50~\cite{he2016deep} architecture with feature output directly modified to 64 or 128 for all of our SimCLR results. We add an extra BatchNorm layer to the end of the model, as suggested in~\cite{chen2020big}. 

\vspace{3pt}
\noindent\textbf{Training} Most of our training procedure is taken directly from the original SimCLR paper~\cite{chen2020simple}. We use the multi-crop augmentation procedure originating from~\cite{caron2020unsupervised}, specifically implemented as in~\cite{caron2021emerging}. 

\subsubsection{Autoencoder}

\vspace{3pt}
\noindent\textbf{Training} We train our AutoEncoder with the Adam optimizer using learning rate $3\times 10^{-4}$ and weight decay 0.0 for 100 epochs with cosine learning rate decay and 5 epochs of linear warmup with batch size 512.

\subsection{ImageNet-1k Training}

\noindent\textbf{Transformation} We train the transformation for 80 epochs with the Adam~\cite{kingma2014adam} optimizer. We use learning rate $5\times 10^{-4}$, weight decay $3.0517578125\times 10^{-5}$, cosine annealing learning rate schedule with one cycle~\cite{cosinelr} with linear warmup~\cite{warmup} for 5 epochs, taken from ResNetV1.5~\cite{repo}. At epoch 40 we freeze the BatchNorm statistics. We find empirically that this makes training more stable for smaller embedding sizes. We suspect that there is some configuration of hyperparameters and learning rates where this is not necessary, however we were not able to find it. For normalization methods with no batch statistics (e.g. LayerNorm~\cite{ba2016layer}), we find that this is not necessary, but we get slightly worse performance (61.2\% vs 61.6\% CMC top-1 in the no side-information case with LayerNorm instead of BatchNorm).

\vspace{3pt}
\noindent\textbf{Old and New Embedding Models} We train the old and new embedding models (standard ResNet50 with 128 dimension embedding) with ResNetV1.5 hyperparameters~\cite{repo}. We train with batch size 1024, learning rate 1.024, weight decay $3.0517578125\times 10^{-5}$, momentum 0.875, and cosine learning rate decay with 5 epochs of linear warmup for 100 total epochs. 

\subsection{Places-365 Training}

\noindent\textbf{Transformation} We train the Transformation for the same duration and with the same hyperparameters as for ImageNet. 

\vspace{3pt}
\noindent\textbf{Old and New Embedding Models} We train the old and new embedding models for the same duration and with the same hyperparameters as for ImageNet. We use embedding dimension 512 for our ResNet50 models. We explain this choice in Appendix~\ref{sec:embed-dim}.

\subsection{VGGFace2 Training}

\noindent\textbf{Transformation} We train the Transformation for the same duration and with the same hyperparameters as for ImageNet. Since the embeddings are normalized in this instance, we normalize the outputs of the Transformation during both training and inference.

\vspace{3pt}
\noindent\textbf{Old and New Embedding Models} We train VGGFace2 with the ArcFace~\cite{deng2019arcface} loss function. Following~\cite{deng2019arcface}, we use a margin of 0.5 and scale of 64, with embedding dimension 128 (we find this to perform equally to embedding dimension 512, which was used in \cite{deng2019arcface}). We also resize to 3x224x224 as we find this does better than 3x112x112, which is a standard for face retrieval applications. We train the old and new embedding models for the same duration and with the same hyperparameters as for ImageNet.

\subsection{BCT Modifications}

\noindent\textbf{Old and New Embedding Models} We use the author provided code~\cite{repoBCT} with a few modifications. In particular, we add the ResNetV1.5 parameters (stated previously) to properly compare with our method and modify the output embedding dimension to 128 (for ImageNet and VGGFace2) and 512 (for Places-182 and Places-365). We find that simply projecting the output layer performs equally well. These modifications result in significant improvement over the original code provided. In particular, $\phi_{new}^{BCT} / \phi_{new}^{BCT}$ performance goes from 60.3\% CMC top-1 before our modifications to 62.4\% CMC top-1 after our modifications.

\section{Transformation Size Ablation}\label{sec:transform-size}

\begin{table}[t]
\centering
\resizebox{0.7\columnwidth}{!}{\begin{tabular}{cc}
  \toprule
\# of params (M) & 
\begin{tabular}{c}
     $(\phi_{new} / h(\phi_{old}, \psi))$  \\
CMC top-1 | top-5 \%
\end{tabular} 
\\
\toprule
0.79 &  62.9 | 81.0\\
1.9 &  64.1 | 81.8\\
5.7 &  65.0 | 82.3\\
19.6 &  65.0 | 82.5\\
\bottomrule
  \end{tabular}}
\caption{Effect of transformation function capacity on accuracy. Accuracy seems to saturate at a relatively small number of parameters. Note that there is not a one-to-one comparison between number of parameters and FLOPS, as convolutional layers tend to have a higher FLOPS to parameter count ratio. See \autoref{fig:costs} for a direct comparison of these attributes.}
  \label{tab:transfer_capacity}
\end{table}
\begin{table}[t]
\centering

\resizebox{0.7\columnwidth}{!}{\begin{tabular}{cc}
  \toprule
Loss & 

\begin{tabular}{c}
     $(\phi_{new} / h(\phi_{old}, \psi))$  \\
CMC top-1 | top-5 \%
\end{tabular} 
\\
  \toprule
MSE &  65.0 | 82.3\\
KL &  60.7 | 78.3\\
KL-Reversed &  55.7 | 76.1\\
\bottomrule
  \end{tabular}}
\caption{Effect of loss function on training of \fct{} transformation.}
  \label{tab:transfer_loss}
\end{table}

\subsection{Transformation Capacity and Training}
The transformation function $h$ should have small memory foot-print and computational cost. In \autoref{tab:transfer_capacity}, for the same setup as ImageNet experiment as in Section~\ref{sec:results}, we show accuracy of the transformed features for a different transformation model architectures with growing width.
In \autoref{tab:transfer_loss} we compare effect of loss function. For KL-divergence, we apply both $\phi_{new}$ and $h(\phi_{old}, \psi)$ to the linear classifier trained with the new model (which is frozen), get log-probabilities, and then apply the Softmax function. We also considered reversed KL-Divergence. Empirically, MSE with target feature outperforms other choices. This has also been observed recently in \cite{budnik2021asymmetric}.

\section{Embedding Dimension ablation}\label{sec:embed-dim}

In this section, we report the effect of embedding dimension in performance of embedding models for ImageNet, Places-365, and VGGFace2. The old and new embedding models' top-1 retrieval performance is shown for different embedding dimensions in \autoref{tab:emb-dim-ablation}. In all cases we directly modify the feature layer output size, rather than projecting the original higher dimension output (e.g., 2048-dimensionsonal features of ResNet50) to a lower dimension, as in~\cite{shen2020towards}. Empirically, for ImageNet-1k and VGGFace2-8631 embedding dimension of 128 obtains highest accuracy, while for Places-365, an embedding dimension 512 performs the best. Interestingly, old model performance is superior to new model performance for Places-365 at embedding dimensions 128 and 256, validating the notion that we need a higher embedding dimension for that particular dataset.

\section{Compressed SimCLR results}\label{sec:genclr}
\begin{figure}[t]
    \centering
    \includegraphics[width=\columnwidth]{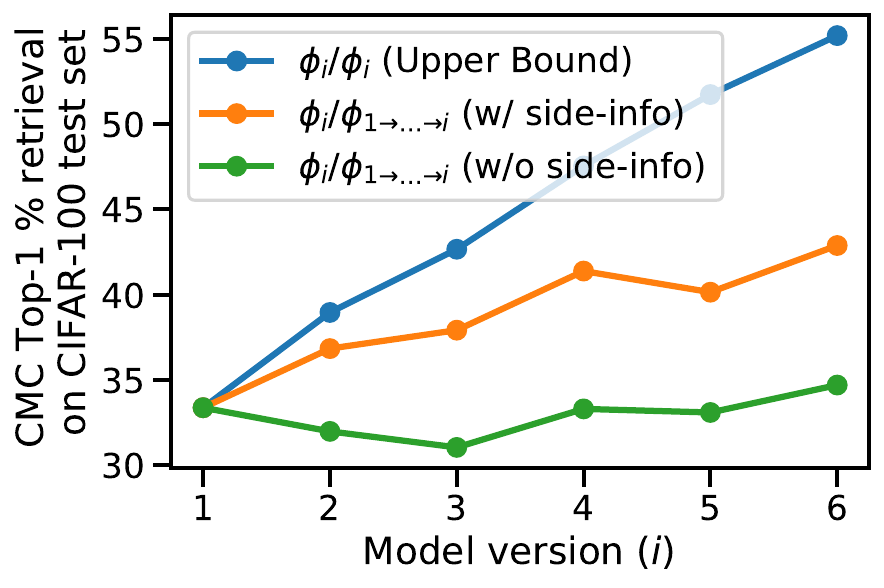}
    \caption{Sequence of updates between models trained on subsets of CIFAR-100 dataset with 50, 60, 70, 80, 90, and 100 classes.}
    \label{fig:c100}
\end{figure}
\cite{ramapuram2021stochastic} presents a method which results in a highly compressed contrastive representation based on SimCLR with similar performance. We used the method in~\cite{ramapuram2021stochastic} to train a SimCLR model with feature dimension $\in \mathbb{Z}_2^{128}$ on ImageNet-500 to be used as side-information. This is a similar setup as in \autoref{tab:imagenet_result}, but with a 32 times more compressed side-information feature vector. We report the numbers for this case in \autoref{tab:compressed}. We see slightly worse performance than with standard SimCLR (-0.9\%), however this shows that side-information representations can be compressed while still maintaining favorable transformation properties.
\begin{table}[H]
\resizebox{0.9\columnwidth}{!}{\begin{tabular}{@{}ccc@{}}
\toprule
Case                                        & CMC top-1|5 (\%) & mAP@1.0 \\ \midrule
$h(\phi_{old}, \psi) / h(\phi_{old}, \psi)$ & 57.2 | 74.7      & 40.0    \\
$\phi_{new} / h(\phi_{old}, \psi)$          & 64.1 | 82.0      & 42.7    \\ \bottomrule
\end{tabular}}
\caption{Results for the ImageNet retrieval setup (see Section~\ref{sec:results}) with a highly compressed SimCLR representation~\cite{ramapuram2021stochastic}.}\label{tab:compressed}
\end{table}

\section{Long Sequence of Small Updates}\label{sec:long}

\edits{In Figure~\ref{fig:c100}, we demonstrate the efficacy of our method for a series of small updates. Model $\phi_i$ is a ResNet50-128 trained on CIFAR-$(10i + 40)$, where CIFAR-$N$ is a subset of CIFAR-100 including only the first $N$ classes. Following our notation from Section~\ref{sec:sequence}, $\phi_{i}/\phi_{1\to\ldots\to i}$ means the query embedding uses $\phi_i$ for indexing and the gallery embedding is obtained with the chain of transformations $h_{i-1\to i}\circ h_{i-2\to i-1} \ldots\circ h_{1\to 2}(\phi_1, \psi_1)$, where each transformation $h_{i-1\to i}$ uses side-information $\psi_{1\to \ldots \to i-1}$. We show that we can achieve meaningful improvement in performance even after the model has drifted quite significantly from $\phi_1$ through a chain of updates. Further, the importance of side-information for sequence is evidence, as without side-information, we sometimes fall short of backward compatibility as defined by~\cite{shen2020towards}.}

\begin{table}[t]
\resizebox{0.9\columnwidth}{!}{\begin{tabular}{@{}ccc@{}}
\toprule
Case                                        & CMC top-1|5 (\%) & mAP@1.0 \\ \midrule
$\phi_{old} / \phi_{old}$                   &  21.9 | 46.8   & 7.09         \\ 
 $\phi_{new} / \phi_{old}$ & 0.3 | 1.5 & 0.12\\
$\phi_{new} / \phi_{new}$                   & 37.0 | 65.1      & 17.0    \\ \midrule
$h(\phi_{old}, \psi) / h(\phi_{old}, \psi)$ & 31.7 | 59.5      & 16.1    \\
$\phi_{new} / h(\phi_{old})$                & 33.3 | 62.3      & 16.9    \\
$\phi_{new} / h(\phi_{old}, \psi)$          & 35.1 | 63.7      & 17.5      \\ \bottomrule
\end{tabular}}
\caption{\fct{} compatibility results when the old and new architectures are trained on completely disjoint objectives. $\phi_{old}$ is a ResNet50-128 trained on ImageNet-1k and $\phi_{new}$ is a ResNet50-512 trained on Places-365. Side-information is SimCLR trained on ImageNet-1k. \fct{} is able to get very close to the upper bound even in this challenging scenario.}\label{tab:disjoint}
\end{table}

\section{ImageNet-1k to Places-365 evaluated on Places-365}\label{sec:disjoint}
\edits{In this scenario, the training objectives of $\phi_{old}$ and $\phi_{new}$ are completely disjoint. This means they share no training data in common. We train $\phi_{old}$ on ImageNet-1k and $\phi_{new}$ on Places-365. For side-information, we used SimCLR trained on ImageNet-1k. We use the Places-365 validation set as both the query and gallery sets for retrieval evaluation. We see that \fct{} is quite successful in this instance with full results presented in Table~\ref{tab:disjoint}. We are able to maintain good model compatibility across different training objectives.}

\section{More Comparisons to Related Methods}\label{sec:lce-rbt}

\edits{In the absence of provided code, we have reimplemented~\cite{meng2021learning} and \cite{wang2020unified} and modified their use for our setting. For~\cite{meng2021learning}, we modify the embedding dimensions to 128, change the loss function to cross entropy instead of ArcFace, and only train the forward transformation, but otherwise keep the original hyperparameters. For~\cite{wang2020unified}, we use their residual bottleneck transformation (RBT) architecture for forward transformation instead of our MLP. For old to new transform performance, RBT achieves $34.0\%$ CMC top-1 ($\phi_{new} / h(\phi_{old})$ in our table) while LCE achieves $56.5\%$.}
\section{Licenses}\label{sec:resources}

\subsection{Software}

\noindent\textbf{PyTorch}~\cite{paszke2019pytorch} is under the BSD license.

\vspace{3pt}
\noindent\textbf{Python} is under a Python foundation license (PSF)

\subsection{Datasets}

\noindent\textbf{ImageNet}~\cite{imagenet} has no license attached.

\vspace{3pt}
\noindent\textbf{Places-365}~\cite{places365} is under the Creative Commons (CC BY) license.

\vspace{3pt}
\noindent\textbf{VGGFace2}~\cite{cao2018vggface2} has no license attached.

\end{document}